# A Concept-Value Network as a Brain Model

Kieran Greer, Distributed Computing Systems, Belfast, UK.
http://distributedcomputingsystems.co.uk
Version 2.3

***Abstract -*** This paper suggests a statistical framework for describing the relations between the physical and conceptual entities of a brain-like model. Features and concept instances are put into context, where the paper suggests that features may be the electrical wiring, although chemical connections are also possible. With this idea, the actual length of the connection is important, because it is related to firing rates and neuron synchronization, but the signal type is less important. The paper then suggests that concepts are neuron groups that link feature sets and concept instances are determined by chemical signals from those groups. Therefore, features become the static horizontal framework of the neural system and concepts are vertically interconnected combinations of these. With regards to functionality, the neuron is then considered to be functional and the more horizontal memory structures can even be glial. This would also suggest that features can be distributed entities and not concentrated to a single area. Another aspect could be signal 'breaks' that compartmentalise a pattern and may help with neural binding.

*Keywords:* feature, concept, electrical-chemical network, binding.

## 1 Introduction

This paper suggests a statistical framework for describing the relations between the physical and conceptual entities of a brain-like model. Features are considered to be horizontal structures that are linked together in a more vertical manner by concepts. The paper suggests that features are in fact the wiring[1], where electrical wiring was the original idea, but chemical wiring is also possible. With this idea, the actual length of the connection is important, because it now determines the firing rate and neuron synchronisation, whereas the actual signal type is less important. The paper then suggests that the concepts are neuron groups that link features through chemical signals and concept instances are dependent on the signal types from those groups. The concept output can vary based on signal composition or strength, for example. Therefore, features become the static framework of the interconnected neural system and concepts are combinations of these. This view of the network would equally relate to a grid, mesh, or tree structure, where features extend horizontally and the concepts vertically. Some test results help to support the theory. The model resembles real neurons mostly in the structural framework, not in how they work internally. So, input and output is assumed to pass through automatically, without dendrites or synapses, for example. But the traditional artificial neural network model of fully-linked layers has been replaced by a much more organic one. It is interesting to think that a real biological model could use a similar design.

[1] Unfortunately, the author has used the term 'synapse' to mean the whole connection between neurons in other papers, but here it more specifically means the synaptic gap. The context will make it clear.

The idea of simply defining the neuron connections as features has come directly from a cognitive model that the author is developing ([8][9][10][12] and related papers), where tests have simply concluded that the statistical processes are plausible. There could be practical reasons for using this model. Apart from energy efficiency, a computer model may find it attractive because it helps to compartmentalise the network. It is also an unsupervised method if the categories are not known. A program would ultimately want to add symbols to the neurons so that everything can be understood. If this is not possible, a defined distance can be used to assign a weak type of symbol to a set of connections, as something that is statistically measurable. This would help to increase the knowledge level by a small amount.

The rest of this paper is organised as follows: section 2 describes some related work, mostly from the biological sector. Section 3 introduces the network idea of features and concepts and makes biological comparisons. Section 4 describes a test result that would support the theory, while section 5 gives some conclusions on the work.

## 2 Related Work

The paper [1] has studied horizontal connections in the cortex, where the arrangement is thought to be mostly columnar. They note that local horizontal connections drop significantly with distance from the cell, but that a lot of non-local (from the sub-cortical area) connections exist and they can serve as a substrate for reliable and temporally precise signals. They conclude that the horizontal connections can help with rapid sensory-motor transformation and decision making, that would not be possible with local connections alone. There is no mention of electrical signals, but clearly the idea is to synchronise patterns that should fire at the same time. The paper [5] states that the most hotly debated question in Neuroscience during the 20th century was whether synaptic transmission, which is the currency of the brain, was mediated electrically or chemically. While chemical activity is preferred, in recent years, electrical transmission has regained recognition and relevance. They note two types of transmission - through pathways or radially through an electrical field. They note different modalities that are quite specific, where different cells have their own specialised ways of transmitting signals. The new framework can therefore only be taken in a very general sense. But the advantage of speed through the electrical signal is clear. They also note the existence of electrochemical transmitters.

The paper [4] describes that some neurons have physically continuous, pipelike connections to other cells, which scientists call electrical synapses. The paper notes that these synapses appear to play a role in the synchronous firing of large sets of neurons. They are however, concentrated in certain brain regions, where synchronisation is critical. The paper [18] describes that chemical and electrical synaptic transmissions closely interact and that both forms of communication might be required for normal brain development and function, even in mammals. It notes that the speed and reliability of electrical transmission can be combined with the ability to induce plastic changes that is characteristic of chemical transmission. It also makes the point that the presence of these two modalities of synaptic transmission does not exclude the possibility that brain cells could also communicate via alternative mechanisms, such as volume transmission (diffusion through the extracellular space of neurotransmitters that reach remote target cells) and by generating electrical fields. This may be important when considering the open question posed in section 3.2. The paper [22] ran some experiments on models that used hybrid chemical-electrical synapses, to investigate the emergence of global and collective dynamic states. They showed that when using hybrid synapses, more dynamic properties emerged in the resulting networks and for large enough electrical synapse coupling, the whole neural networks became synchronised. Their excitatory population was also connected through small-world topology [21]. They showed that the electrical synapses helped to synchronize firing but may also mediate spiking among neuron clusters. They also noted that neurons with electrical connections are spatially very close. Two papers [16][2] measured neuron synchrony and activation. The paper [16] built a detailed computational model of the brain and noted that synchronisation and oscillation between the neurons is related to distance, or a small delay in firing sequences, more than signal strength or decay time constant. Electrical synapses were included in the model, where they found that the power of the network oscillations increased in a non-frequency-selective manner when adjacent neurons were connected by purely resistive electrical synapses. They concluded that delay-induced synchronization should be considered for fast oscillations in particular (>40Hz), but also that some types of fast oscillations appear to persist, even in the absence of synaptic transmission (200 Hz 'ripples'), and axoaxonal gap junctions may be essential for their generation. The paper [20] also considers this problem.

Axo-axonic synapses [3][16] consist of an axon terminating on another axon and therefore do not require action potential activity. They are in fact one of 3 types, where the others are axon to dendrites or axon to soma. As written in [3], several studies demonstrate that axon terminals synapsing on another axon terminal provide direct neural circuit routes for presynaptic modulation. They may exert rapid and functionally significant events, without temporal or spatial summation. Neuromodulation can then change neurons and neural ensembles, to prepare them for firing. The paper [17] states that neuromodulators target ion channels and synaptic interactions to modify circuit dynamics, which allows for adaptability of circuit operation in different behavioral contexts. Synaptic modulation is not limited to changes in the strength of connections, but involves modifications of short- and long-term synaptic plasticity. Neuromodulators can act globally in a behavoural sense, or locally on any single neuron and therefore, reconfiguration of neural circuits by neuromodulators is an intricately balanced process.

The paper [2] measured the effects of a single neuron and found that it would excite its immediate neighbourhood, but then switch off most of the area outside of that. It would, in fact, compete for the input, rather like a Self-Organising Map [15] and is described in terms of feature competition and amplification. This would help to reduce noise and create a clearly defined cluster. Then some neurons further away, also tuned to the input, would also get excited. However, neurons that were tuned to respond to similar features competed and more strongly suppressed each other, compared to neurons with a different tuning. This inverse relationship remained true regardless of the wiring distance between neurons and suppressing the immediate surrounding area would give the pattern more definition. 'The response patterns of these neurons were also well correlated in time with that of the target neuron (that is, their moment-to-moment electrical activities closely resembled each other).' The message here is that a radial distance from a neuron is more important than link lengths and it shows that a single neuron can have a significant effect on the whole network. But it also depended on neurons that were tuned to each other, so there is some consistency with the feature network of this paper, which would also turn a single neuron into a small network of neurons.

## 3 Features and Concepts

With this design, features are the static structure of the network and concepts are instantiations of feature sets. Features and concepts represent fundamentally different things, even though they are integrated in the same structure. The features are static arrangements while the concepts are value based and dynamic. A feature is an arrangement of the same, while a concept is an aggregation of different parts. This view of the structure would also relate to a grid, mesh, or tree structure, where features extend horizontally and the concepts vertically. Considering the network structure then, it may be the case that the between-neuron distances are significant and can be compared to features. This makes good sense, because if equidistant neurons fire at the same time, then they are likely to represent the same thing and can be mapped to the wiring itself. It is also suggested that this wiring may be electrical, because it wants to fire immediately. Electrical paths have been shown to be bi-directional, which would be an advantage. If the purpose is simply to activate all patterns that contain the concept, then the signal can travel in either direction to any other neuron in the feature set. It will also occur quickly, because the electrical signals travel faster and so the active regions can all fire at the same time. However, the discovery of axoaxonal synapses is also interesting and chemical signals can also be sent horizontally. In section 3.2, glial cells will even be suggested. A small point may be the following: it was found to be mathematically the case that equal spacing between neurons is the most economic setup, with regards to energy usage [12]. Therefore, if a set of neurons are firing together as a single concept, equal spacing between them would be best. The paper also suggested that when adding new neurons, halving the between-neuron distance each time is the most economic. A feature length may therefore not be completely variable, but may be graded. Links to other firing features should be less numerous, but they would similarly synchronise as part of their own network, maybe using a small-world effect. Then, the closely connected neurons would possibly reinforce each other, but if other areas further away are also part of the feature and are tuned to it, they can get activated as well.

### 3.1 Integrated Model

The paper [8] re-defines the 3-level architecture into a more human-like vernacular. The upper cognitive layer is not considered yet and so there is a mapping for the first two levels only. These are a bottom optimising layer and a middle aggregation layer. Considering the bottom optimising layer, while it is for optimising links in patterns, it is also described in terms of 'Find' or 'What' functionality. This has an obvious mapping to features, because any search process starts with a set of initial features. It is also horizontal in nature, because the search will try to find best combinations of the feature set before moving to the next stage. The middle layer is for aggregation of the links which also includes averaged decisions over them. As the middle layer nodes would receive feature sets as input, they can be aggregators of features into concepts, or concepts into larger concepts, for example. This can relate to a 'Compare' or 'Why' function over signal values. Aggregation through the nodes is therefore more vertical in nature. The scenario is therefore chains of electrically connected nodes, extending themselves through chemical connections, to link with other chains, where the whole structure is a concept.

In Figure 1 for example, there is a 'tree' feature that is distributed. If the between-node lengths are the same, then they should synchronize and support each other. If there is a brain search about hobbies, then that signal type is sent to the tree feature ('Find'). One tree concept instance extends into the rainforests domain and another extends into arts and crafts ('Compare'), and will activate depending on which signals are sent. Then if there are other criteria from other places, where they converge can determine the result. The vertical concepts can remain distinct, but are also joined by the base 'tree' concept.

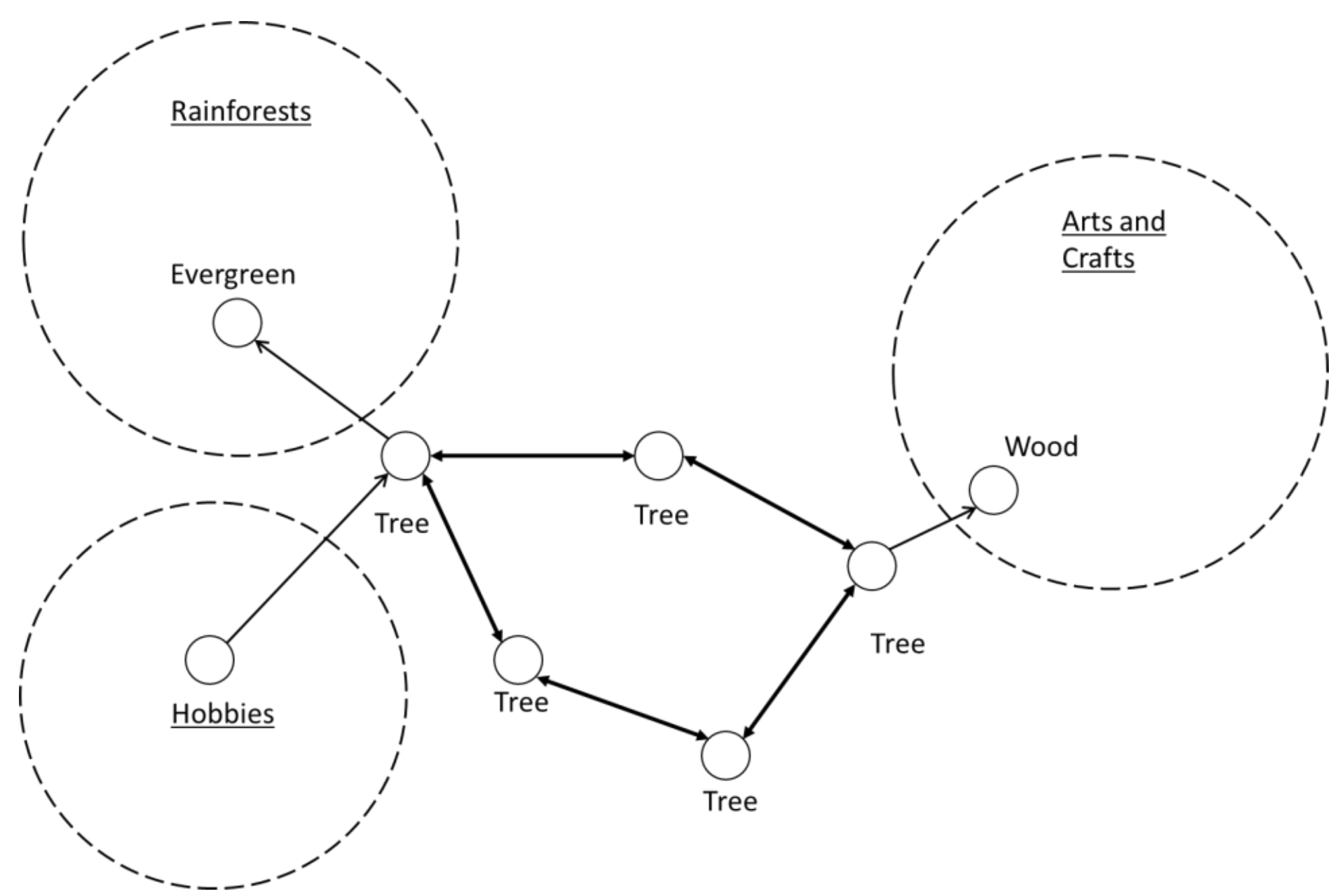

Figure 1. A distributed 'Tree' feature with specific concept instances to different domains.

### 3.2 Memory and Function

The author is writing a larger test program for a whole cognitive model. Due to resource constraints, most of the nodes in it are modelled using a simpler class, that represents glial cells instead. The neuron class with an axon and a dendrites, has not been used very much. This really supports the idea that neurons may be functional and that most static memory structures can be modelled using a simpler type of cell. Therefore, the functionality that is more vertical through the cells is neural and the horizontal memory structures can be glial. If this is the case, then glial cells, electrical wiring and axo-axonic wiring can all perform a similar task. Modern research notes that glial cells may be active in modulating neuronal signals and so this is an interesting possibility. Simply activating a signal in different places does not require a lot of functionality, but it would have to transfer all the required information for the concept instances to be consistent. This is therefore still an open question, but it is also a question for neural models in general. How is the same signal propagated to every node in a pattern, if signal transmission takes place only through dendrites and axons?

Modern research [13] also suggests that the ratio of glial cells to neurons is close to 1:1. This may be interesting when thinking about Shannon's Information Theory [19]. He considered the amount of choice when using the English language and concluded that the redundancy of ordinary English is roughly 50%. This means that when we write English half of what we write is determined by the structure of the language and half is chosen freely. This is therefore consistent with the idea that more static structure determines about 50% of what we do and so it might be represented by 50% of the human brain. The other 50% is more free and value-based. As in [7], brain function expresses, or partly represents, the brain structure.

### 3.3 Neural Binding

As stated in [10]: at its most basic, Neural Binding [6] means ‘how do neural ensembles that fire together be understood to represent the said concept’. It includes the idea of consciousness and how the brain is able to be coherent. Some cognitive models for the real brain include temporal logic to explain how variables can bind with each other and this includes the flow of information in both directions. Neural Binding comprises at least 4 distinct problems, some of which remain unsolved. The famous ‘red square, blue circle’ problem usually relates to how the brain can process and understand visual input. As stated in [6]:

> 'The basic question on visual feature-binding is ancient – why don't we confuse, e.g., a red circle and a blue square with a blue circle and a red square. While linking features to the correct object and location is a requirement for effective vision it is not normally a problem, in the sense of being mysterious. The visual system is spatiotopically organized and most detailed vision is done in foveal fixations which are inherently coordinated in space and time.'

The discussion for this paper is more interested in the variable binding problem, which remains unsolved. Rather than recognising a red square and blue circle from visual input, these objects would be generated internally in the human mind. Thus, there must be pattern representations of them in the brain, but the problem is – how can the individual pattern concepts get shared, while these pattern constellations remain distinct? As discussed in [6], one option called phase binding breaks the cycle of neural firing into discrete time slices, which could divide local firing patterns into separate phases, but the idea is contentious. Temporal phase coherence is no longer considered a major contender in feature binding, in part because it would be much too slow. It is much more relevant in variable binding where most other models don't apply. The other basic model is brute force enumeration of all possible variable bindings. The problem is that there are a potentially unbounded number of items that might be bound to a variable, so none of the pair-coding techniques can work.

This paper can therefore make a new suggestion that is compatible with a concept-value network. Section 3.1 describes that the electrical, or possibly glial, signals would probably not convey any chemical signal that they continued from. But could this break in the signal transmission also be helpful? Figure 2 gives just one possible scenario for linking the 'red square, blue circle' concepts. Consider that (chemical) signals travel from the bottom row to the top row. In order to complete a circuit however, the concepts also need to be connected outside of this signal transmission. It is interesting that one circuit travels in a clockwise direction, while the other circuit travels anti-clockwise. The main point however is that some of the connections that complete a circuit may not have to carry a (chemical) signal and could in fact act as signal breaks. This would, in effect, compartmentalise the pattern group, but also slightly dynamically. This could be a very helpful feature of the architecture, where these natural breaks would help to define what pattern concepts occur together and therefore make them distinct.

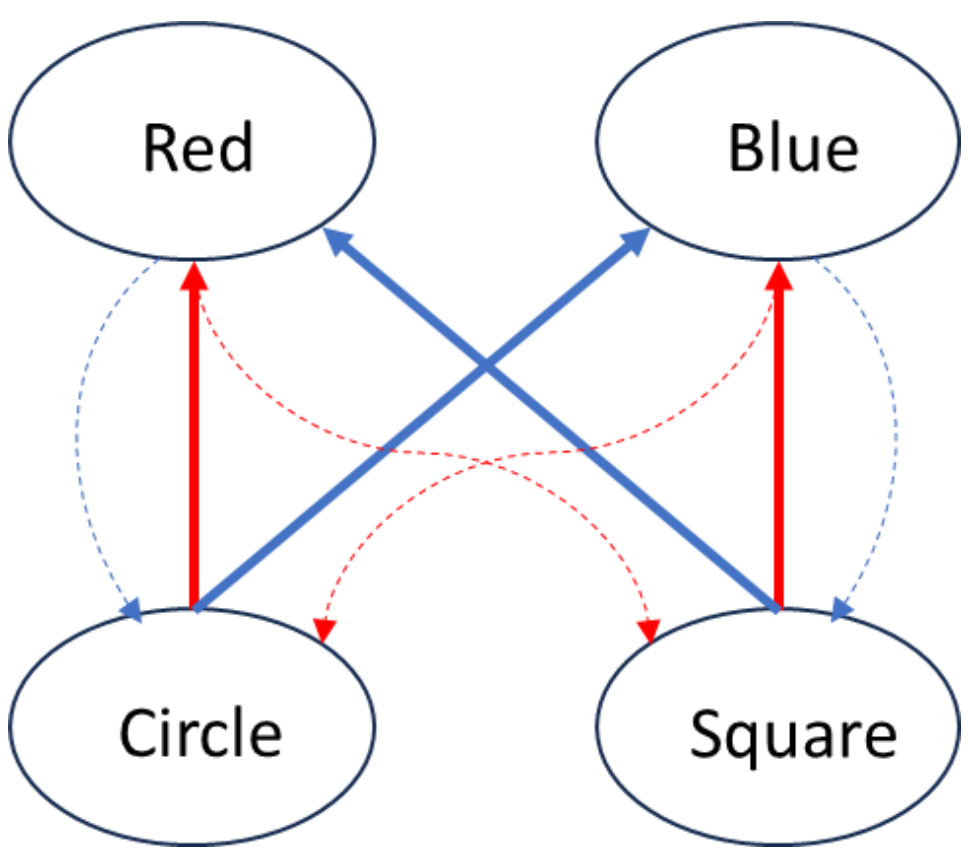

Figure 2. Solid lines indicate connections with a (chemical) signal and dashed lines without. The same colour completes a circuit, where the 2 circuits travel in opposite directions.

## 4 Feature – Value Test

Some of the earlier research has realised a number of classifiers that can be used to cluster data. For example, the cohesion equation of [8] was compared to the Chi-Square measure [14] in the paper, but as it deals more with subsets of data rows and not columns, it probably measures something different. Both are entropic, but Chi-Square measures a goodness of fit for independent variables, while the cohesion equation was a measure of similarity. These 2 equations therefore showed the contrast that would be expected. In a new set of tests, a computer program presented parts of an ontology to other clustering algorithms, to see how well they would re-construct the ontology. A level of noise was set, but if this is very low, then the algorithms would be expected to re-construct the ontology exactly. Two different types of clustering mechanism were used, which was the linking mechanism [11] and the Frequency Grid [9]. In this case, the Frequency Grid is more feature-based and again entropic, while the dynamic linking mechanism is more value-based and local. The algorithms learned the small cluster parts of the ontology, as determined by the random presentations, but to learn every link would require a lot more iterations. The two cluster sets were slightly different however and if they were then combined, it gave a much better view of the whole cluster sets that they were learned from. Because there was overlap in the two cluster sets, putting them together would actually give a better view of the whole ontology.

As an example, an ontology was written with 4 different entity types, each with 10 instances, giving a total of 40 nodes. There were also 6 inter-pattern links to count as noise. A random number of up to 5 nodes from a pattern would be presented each time, where 50% of the time an inter-pattern link would be selected if it existed, and the clustering algorithms would learn the correct associations. After 500 iterations, the linking mechanism and the frequency grid had learned the following information about the underlying ontology: The linking mechanism created 10 clusters and the frequency grid created 16 clusters, with an inherent problem that some clusters have only 1 entry. Because of the overlap, if the two cluster sets were combined, the original 4 cluster sets would be realised. Thus, there is a question about whether these two views can give a better picture of the whole, or if they can re-construct the whole in a quicker time.

## 5 Conclusions

This paper gives one possible view of a brain-like structure that may be biologically plausible. The framework has been reduced to the bare components that interact through statistics only. If the network wiring stores the features, then an input stimulus activating different sets of neurons over this can be flexible enough to create different types of concept, from the same feature sets. Tests showed that two different clustering mechanisms, one representing the environment features and one representing the environment instances, produced a much better cluster description when combined than separately. Each produced small sets of clusters that overlapped and if the overlap was also considered, then the cumulated cluster was a much better representation of the whole entity. A reason for both electrical and chemical connections can be made by the model. The electrical wiring can connect neurons horizontally, when they represent the same thing. It can fire automatically because of this and can be quicker, because these same components want to be active at the same time. It can oscillate because the direction is both ways, to ensure that all neurons in the network path are always active.

With regards to functionality, the neuron is then considered to be functional and the more horizontal memory structures can even be glial. Then the chemical wiring can be variable, with regards to its signal and therefore it can change the values of the neurons and therefore also their meaning. This is turn means that the neurons are capable of storing and reacting to more than 1 value type, but there is an open question here. If the feature is activated horizontally across nodes, then is the original signal transferred with it? Answering this for activating patterns in general should solve it, or things like firing rate and strength may become more important than chemical composition. There may also be an advantage to not propagating the chemical signal every time, which could help with the neural binding problem.

If thinking about how the brain incorporates new information, it might actually be easier to create another new small tree to represent a new concept and link it with the existing tree, instead of trying to merge the two patterns in some precise way. This would include for base neurons firing close to each other, or some distance apart. This could suggest that concepts are in fact distributed entities and not concentrated in a single area. There is also a statistical property involved, in that a base feature with more nodes will be easier to find. More nodes would mean more options from the trees they emanate. In fact, a set of small trees will be more distinct than a single larger tree. Thus, the base concept node would indeed get repeated, but then the other concept information could remain more separate and would also be functionally distinct.